\pdfoutput=1

\documentclass[11pt]{article}

\usepackage[final]{acl}

\usepackage{times}
\usepackage{latexsym}
\usepackage{url, graphicx, multirow, booktabs, subcaption, float, comment}

\usepackage[T1]{fontenc}

\usepackage[utf8]{inputenc}

\usepackage{microtype}

\usepackage{inconsolata}

\title{Can GPT-4 do L2 analytic assessment?}

\author{Stefano Bannò, Hari K. Vydana, Kate M. Knill, Mark J. F. Gales \\
         ALTA Institute, Department of Engineering, University of Cambridge (UK) \\
    \texttt{\{sb2549,hkv21,kmk1001,mjfg100\}@cam.ac.uk}}

\begin{document}
\maketitle
\begin{abstract}
Automated essay scoring (AES) to evaluate second language (L2) proficiency has been a firmly established technology used in educational contexts for decades. Although holistic scoring has seen advancements in AES that match or even exceed human performance, analytic scoring still encounters issues as it inherits flaws and shortcomings from the human scoring process. The recent introduction of large language models presents new opportunities for automating the evaluation of specific aspects of L2 writing proficiency. In this paper, we perform a series of experiments using GPT-4 in a zero-shot fashion on a publicly available dataset annotated with holistic scores based on the Common European Framework of Reference and aim to extract detailed information about their underlying analytic components. We observe significant correlations between the automatically predicted analytic scores and multiple features associated with the individual proficiency components. 
\end{abstract}

\section{Introduction} \label{introduction}

Automated essay scoring (AES) of second language (L2) proficiency is a well-established technology in educational settings, involving the automatic scoring and evaluation of learners’ written productions through computer programs~\cite{shermis2003}.

Originating in the 1960s, the roots of AES can be traced back to the development of Project Essay Grade (PEG)~\cite{page1966imminence, page1968}, an automatic system which evaluated writing skills based only on proxy traits: hand-written texts had to be manually entered into a computer, and a scoring algorithm then quantified superficial linguistic features, such as
essay length, average word length, count of punctuation, count of pronouns and prepositions,
etc. Across the following decades, as natural language processing (NLP) technologies have advanced and increased their power~\cite{landauer2003}, the field of AES has expanded and improved, and more significant studies have been conducted from the 1990s and early 2000s. The most widely known automated scoring systems for essays include the e-rater\textregistered, developed by Educational Testing Service \citep{burstein2002erater, attali2006automated}, IntelliMetric{\texttrademark} by Vantage Learning~\citep{rudner2006evaluation}, and the Intelligent Essay Assessor\texttrademark, built at Pearson Knowledge Technologies~\citep{landauer2002intell}.

In recent years, deep neural network (DNN) approaches have brought significant improvements~\cite{alikaniotis2016automatic}, and especially the advent of transformer-based architectures~\cite{vaswani2017attention}, such as BERT~\cite{devlin2018} which took the world of NLP and, consequently, AES by storm, outperforming classic feature-based systems~\cite{rodriguez2019language}. Yet, the most recent breakthrough has been brought by large language models (LLMs), such as the GPT models~\cite{brown2020language, openai2023gpt4}, which might revolutionise the world of AES, not only from the NLP experts' and language testers' perspective, but also considering the users' point of view due to GPT's extremely accessible and intuitive interface. In the context of L2 writing assessment, previous studies have employed GPT-3.5~\cite{mizumoto2023gpt} and GPT-4~\cite{yancey2023rating}, obtaining promising results.

Although LLMs have been employed for holistic scoring (i.e., assessing the overall quality of a composition as a whole, considering various aspects such as vocabulary, grammar, coherence, etc. altogether), to the best of our knowledge, so far they have not been investigated for the task of analytic scoring (i.e., breaking down a composition into specific components or criteria and assigning separate scores or ratings to each component).\footnote{\citet{naismith-etal-2023-automated} investigated the use of GPT-4 on a proprietary dataset annotated with specific scores targeting coherence only.} Offering L2 learners specific analytic proficiency scores is crucial for delivering insightful and effective feedback, emphasising both their strengths and weaknesses to facilitate improvement.

For holistic scoring, previous works have shown that state-of-the-art automatic techniques can reach near-human results~\cite{alikaniotis2016automatic, taghipour-ng-2016-neural} or even outperform them~\cite{rodriguez2019language}. This is, at least in part, ascribable to the fact that holistic scores are generally easier to obtain for human evaluators (see Section \ref{section2}). Conversely, assessing analytic aspects of language proficiency is generally considered to be more difficult, time-consuming, and cognitively demanding for human evaluators, and, as a result, ``noisy'' ground truth scores are harder to learn and predict for automatic systems (see Section \ref{section2}).

Starting from these premises, in this paper, we conduct a series of exploratory experiments on a publicly available dataset annotated with holistic scores according to the Common European Framework of Reference (CEFR)~\cite{cefr2001, cefr2020} using GPT-4 in a zero-shot fashion, and aim to extract specific information about their underlying analytic components. Although ground truth analytic scores are not available, we find significant correlations between the analytic scores predicted by the model and several features related to the analytic scores.

\section{Holistic versus analytic scoring} \label{section2}

\subsection{Human assessment}

Holistic and analytic approaches to assessing L2 proficiency are commonly utilised, differing in scoring methods, underlying assumptions, and practical application. While holistic assessment consists of assigning a single overall numerical score to a specific performance based on a singular set of rating criteria, analytic assessment involves providing various sub-scores to the performance based on multiple sets of criteria. As a result, there are conceptual differences between the two approaches~\cite{barkaoui2011}. Holistic assessment typically assumes that the construct being evaluated is a unitary entity and can be represented on a single scale. While this approach acknowledges that the construct may consist of various elements, it implies that development across various aspects of proficiency is uniform. Conversely, analytic assessment views the construct as multi-dimensional and advocates for a multi-faceted assessment, recognising that development across various aspects may be irregular. For instance, the levels of the CEFR are structured according to `can-do' descriptors of language proficiency outcomes and expect evaluators to grade proficiency by means of holistic assessments. Nonetheless, the CEFR levels do have a modularisable structure with multiple underlying components (e.g., vocabulary range, vocabulary control, grammatical accuracy, etc.), acknowledging that a learner may be more proficient in certain aspects than others~\cite{cefr2001, cefr2020}.

When we consider assessment strictly from a human perspective, holistic assessment is considered highly practical as it is more time-efficient per se and in relation to rater training~\cite{white1984holisticism}, less cognitively demanding~\cite{xi2007evaluating}, and generally has a higher inter-annotator agreement~\cite{weigle2002assessing} than analytic assessment. On the other hand, holistic scoring may suffer from lack of clarity regarding how different aspects are prioritised, which may vary among evaluators~\cite{weigle2002assessing, xi2007evaluating}, the risk that evaluators might primarily concentrate on candidates' strengths rather than their weaknesses~\cite{bacha2001writing}, and the potentially erroneous assumption that various aspects of proficiency develop uniformly over time~\cite{kroll1990second}.

Analytic assessment allows for a more detailed and systematic evaluation and is supposed to provide much more detailed feedback to L2 learners, by highlighting their fortes and their weaknesses~\cite{hamplyons1995rating} in addition to enhancing scoring validity. However, it is not a panacea. Analytic scores may be psychometrically redundant~\cite{lee2009towards} due to a halo effect~\cite{engelhard1994examining}, whereby raters fail to distinguish between different aspects of learners' performances but assess all or some of them with similar scores. For example, when assessing grammatical accuracy, raters might be influenced by the score previously assigned to vocabulary range. On top of this, raters might confuse analytic criteria in the phase of assessment due to high cognitive load~\cite{underhill1987testing,cai2015weight} or, more simply, to indefiniteness of the analytic criteria~\cite{douglas1997theoretical}. The difficulty in providing analytic scores --- especially for a large number of written productions --- is evident in the total absence of publicly available L2 English learner datasets annotated in this way\footnote{\label{footnote_asap}To the best of our knowledge, the only formerly publicly available dataset annotated with analytic scores is the ASAP dataset (\url{kaggle.com/c/asap-aes/data}), but the test data are no longer available for evaluation and comparison with previous work. Furthermore and most importantly, it contains essays written by L1 English speakers.} and the fact that the primary emphasis in AES research has been on holistic scoring.

\subsection{Automatic assessment}

The introduction of automatic assessment techniques --- and especially their recent advancements --- have started to change the game. For holistic scoring, DNN-based systems reached near-human performances~\cite{alikaniotis2016automatic,taghipour-ng-2016-neural}, and the application of transformers-based architectures even beat human inter-annotator agreement~\cite{rodriguez2019language}. However, a notorious problem lies in the impossibility to enter the black box of neural scoring models, and this poses a challenge for explainability and interpretability of the machine-generated holistic scores. Even more so, it is important to explore the ability of automatic models to evaluate specific aspects of language proficiency through analytic scoring: if it is not possible to decompose the holistic assessment process by peeking inside the black box, it may be possible to reconstruct holistic scores starting from their analytic components (with the caveat that we should keep in mind the potential unreliability of human analytic scores, as discussed above). In this regard, automatic systems have been found to be generally better at evaluating specific linguistic phenomena, whilst humans tend to focus on more general aspects of proficiency. For example, \citet{enright2010} suggested that human raters might achieve higher results when assessing ideas, content, and organisation, whereas automatic systems might have better performances when evaluating microfeatures at the grammatical, syntactic, lexical, and discourse
levels. It should be noted, however, that these limitations attributed to automatic systems may no longer necessarily be true in light of the recent advancements involving neural systems, which can be used quite effectively also to assess higher-level aspects of proficiency. For example, previous studies have focused on specific traits of written productions, such as organisation, content, word choice, sentence fluency, narrativity, etc. \cite{hussein2020trait-based,mathias-bhattacharyya-2020-neural, ridley2021}, but they have used the ASAP dataset, which is problematic for reproducibility 
 and only features essays written by L1 English speakers (see note \ref{footnote_asap}). For L2 speaking assessment, the initial study by \citet{banno2022view} investigated the use of multiple different graders, each of which focused on a different set of features related to a specific proficiency aspect.

The introduction of LLMs could be a further game-changer, considering their outstanding results in a broad range of tasks.

To sum up, given that:

\begin{itemize}
    \item holistic scores are generally easier to obtain both from human and automatic graders and generally have a higher inter-annotator agreement, hence higher reliability;
    \item analytic scores are difficult to obtain and might not always be sufficiently reliable;
    \item more often than not, L2 learner datasets are annotated with holistic scores only;
    \item LLMs have been proven to be extremely powerful tools in many NLP tasks;
\end{itemize}

we pose the following research question: 
\begin{quote}
    is it possible to extract information about analytic aspects from L2 learner essays and their assigned holistic scores using GPT-4?
\end{quote} 

\begin{figure*}[t]
\centering
\includegraphics[scale=0.5]{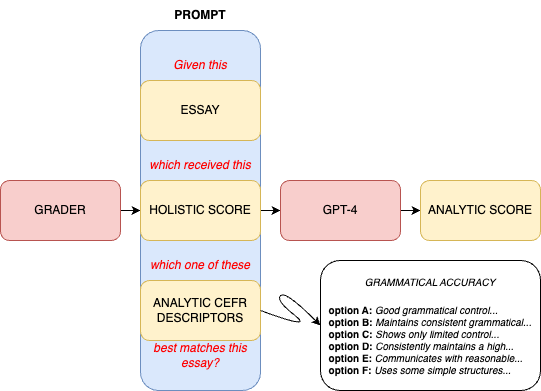}
\caption{The pipeline presented in this study. Grammatical accuracy is only one of the aspects considered.}
\label{pipeline}
\centering
\end{figure*}

Figure \ref{pipeline} shows the pipeline adopted in this study, which will be illustrated in detail in Section \ref{experimental_setup}.

\section{Data}
\subsection{Write \& Improve}

Write \& Improve (W\&I) is an online platform where L2 learners of English can practise their writing skills~\citep{yannakoudakis2018developing}. Users can submit their compositions in response to different prompts, and the W\&I automatic system provides assessment and feedback. Some of these compositions have been manually annotated with CEFR levels and grammatical error corrections since 2014, resulting in a corpus of 3,300 texts, partitioned into a training set of 3,000 and a validation set of 300 essays.\footnote{The dataset can be downloaded from this link: \url{huggingface.co/datasets/wi_locness}.} The proficiency scale ranges from A1 to C2 but also has intermediate levels, resulting in 12 levels, that we arranged on a scale from 1 to 6.5, where 1 is A1, 1.5 is A1+, 2 is A2, 2.5 is A2+, etc., as shown in Table \ref{scores_alignment} (see Appendix \ref{sec:appendix_d}).

\subsection{EFCAMDAT}

Arguably the largest publicly available\footnote{\url{ef-lab.mmll.cam.ac.uk/EFCAMDAT.html}} L2 learner corpus, the second release of EF-Cambridge Open Language Database (EFCAMDAT) \citep{geertzen2013automatic, huang2017ef, huang2018dependency} comprises 1,180,310 scripts written by 174,743 L2 learners as assignments to Englishtown, an online English language school. The compositions are annotated with a score on a scale from 0 to 100 and a proficiency level from 1 to 16 (mapped to CEFR levels from A1 to C2).\footnote{\url{englishlive.ef.com/en/how-it-works/levels-and-certificates/}} In order to align them to the proficiency levels in the W\&I dataset, we normalised the scores as described in Table \ref{scores_alignment} (see Appendix \ref{sec:appendix_d}). For our experiments, we selected a subset of data consisting of 753,508 essays for the training set and 7612 for the validation set, following a similar process to \citet{banno23b_slate}.

\section{Experimental setup}\label{experimental_setup}

\subsection{Longformer-based holistic grader}

Following the pipeline illustrated in Figure \ref{pipeline}, we start our experiments from training a holistic grader, which consists of a Longformer model~\cite{beltagy2020longformer} in the version provided by the HuggingFace Transformer Library,\footnote{\url{huggingface.co/allenai/longformer-base-4096}} a dropout layer, a dense layer of 768
nodes, a dropout layer, another dense layer of 128 nodes, and finally, the output layer. The baseline model (W\&I) is trained on the W\&I training data and optimised on the W\&I validation data using an Adam optimiser~\cite{kingma2014} for 3 epochs with batch size 16, learning rate 1e-6 and mean squared error as loss, but our best-performing model --- which is the one we will use in the following steps of our pipeline --- is trained on the EFCAMDAT training set and optimised on the validation data from the same dataset for 0.5 epochs with batch size 16 and learning rate 1e-5, and subsequently fine-tuned on the W\&I training data and optimised on the W\&I validation data for 4 epochs.

To evaluate the holistic grader performance, we use Pearson's correlation coefficient (PCC), Spearman's rank coefficient (SRC), and root-mean-square error (RMSE).

\subsection{GPT-4-based analytic graders}

Once we obtain the holistic scores from the Longformer-based model, we move on to feeding them into GPT-4 (\emph{``gpt-4-1106-preview''}) to extract analytic scores. Specifically, the analytic scores are related to 9 proficiency aspects as described in \citet{cefr2020}, reported in Appendix \ref{sec:appendix_a}. Five of them compose the linguistic competence: \emph{general linguistic range, vocabulary range, grammatical accuracy, vocabulary control,} and \emph{orthographic control}; while the remaining four form the pragmatic competence: \emph{flexibility, thematic development, coherence and cohesion,} and \emph{propositional precision}.

We excluded sociolinguistic appropriateness because it is not consistently elicited in the W\&I essays, as well as the aspects strictly related to speaking proficiency (i.e., phonological control, turntaking, and fluency) for obvious reasons.

The prompt given to GPT-4 can be found in Appendix \ref{sec:appendix_c}. To exclude potential biases, the holistic scores are fed in their numerical form (i.e., from 1 to 6.5) instead of the original CEFR notation (i.e., from A1 to C2+), and the analytic CEFR descriptors are provided in random order and, obviously, without any reference to the CEFR levels. For completeness, we also try this experiment without giving GPT-4 the holistic score.

At the end of the process, the option selected by GPT-4 is mapped back to its respective CEFR level. 

\subsection{Explanation of the features}

As mentioned in Section \ref{introduction}, the W\&I dataset does not include analytic scores, but we find significant correlations with relevant features extracted from the essays (see Tables \ref{gpt4results_predicted} and \ref{gpt4results_ground}).

\noindent \textbf{\%gram.} refers to the grammatical error rate, which is the number of grammatical error edits divided by the number of words in the essay. These edits are extracted by feeding the original and corrected versions of the W\&I essays into the ERRor ANnotation Toolkit (ERRANT)~\cite{bryant2017automatic}.

\noindent \textbf{\#dif.wds.} is the number of unique difficult words extracted with \texttt{textstat}.\footnote{\url{github.com/textstat/textstat}}

\noindent \textbf{\#unq.wds.} refers to the number of unique words.

\noindent \textbf{\%l.d.t.} is the percentage of text types that are content words obtained using TAACO (Tool for the Automatic Analysis of Text Cohesion) 2.0~\cite{crossley2019tool}.

\noindent \textbf{\#unq.n.chunks} refers to the number of unique noun chunks identified and extracted using spaCy.~\footnote{\url{spacy.io/}}

\noindent \textbf{\#unq.q.m.a.} refers to the number of unique qualifiers, modality markers, and ambiguity indicators identified and extracted using spaCy.

\noindent \textbf{fl.-kinc.} is the Flesch Kincaid readability score~\cite{kincaid1975derivation}, obtained using \texttt{textstat}.

\noindent \textbf{w2v} is the average word2vec~\cite{mikolov2013efficient} similarity score between all adjacent paragraphs, extracted with TAACO 2.0.\footnote{Initially, we also extracted the similarity score using Latent Semantic Analysis~\cite{landauer1998introduction} and Latent Dirichlet Allocation~\cite{blei2003latent}, which showed similar figures, but we did not include them due to reasons of space.}

\noindent \textbf{av.s.ln.} is the average sentence length.

The correlations between these features and the analytic scores are evaluated using SRC since we do not necessarily expect a linear correlation between the two. For example, it is well-known that certain grammatical errors are absent or rare in the A1 level, increase after B1, and then decline again by C2~\cite{hawkins2010}.

\section{Experimental results}

\subsection{Holistic scoring}

Table \ref{longformer_results} shows the results of the Longformer-based holistic graders on the W\&I validation set in terms of PCC, SRC, and RMSE. The model pre-trained on EFCAMDAT and fine-tuned on the W\&I training set outperforms the baseline across all metrics as expected. These results should confirm that holistic grading is a relatively easy task and, since the training data are fully publicly available, potentially within everyone's reach.

\begin{table}[ht!]
    \centering
    \begin{tabular}{c|c|c|c}
    \toprule
       
        \textbf{Model} &  \textbf{PCC} & \textbf{SRC} & \textbf{RMSE} \\
        \hline
        W\&I & 0.707 & 0.772 & 1.267 \\
        EFC+W\&I & 0.866 & 0.874 & 0.786 \\

    \bottomrule
    \end{tabular}
    \caption{Holistic scoring results on W\&I validation set.}
    \label{longformer_results}
\end{table}

\subsection{Holistic score reconstruction}

Once we obtain the holistic scores from the Longformer-based grader, we are ready to feed them into GPT-4. However, before moving on to the analysis of the individual analytic scores, we first calculate the correlation between the average of the predicted analytic scores --- when providing GPT-4 with the holistic scores from the ground truth (GT) or the Longformer-based grader (EFC+W\&I), or with no holistic score (-) --- and the holistic scores, both the ground truth (GT) and the scores automatically predicted by the Longformer-based grader (EFC+W\&I), as shown in Table \ref{holistic_scores_reconstruction}. 
\begin{table}[h]
\centering
\begin{tabular}{c|c|c}
\toprule
\textbf{GPT-4 Prompt} & \multicolumn{2}{c}{\textbf{Reference}} \\
\textbf{Holistic Score} & \textbf{GT} & \;\textbf{EFC+W\&I} \\
\hline
 GT  & \;\;\;\;\;0.904\;\;\;\;\;  & 0.874  \\\hline
\;EFC+W\&I\; & 0.828  & 0.898  \\\hline
 - & 0.797  & 0.827  \\
\bottomrule 
\end{tabular}
\caption{SRC correlation between the average of the predicted analytic scores and the holistic scores.}
\label{holistic_scores_reconstruction}
\end{table}

The first result that catches the eye is that GPT-4 reaches a significant correlation of 0.797 when it is not provided with additional information about holistic scores (-), although this does not necessarily mean that all the underlying analytic scores are effectively targeting their respective proficiency aspects, as we will discuss in the next section. Secondly, it is interesting to observe that the two sources of holistic score in the prompts (i.e., GT and EFC+W\&I) result in the information derived from these scores being used in a non-deterministic fashion, introducing a certain degree of variability.

\subsection{Analytic scoring}

\begin{table*}[t!]
\footnotesize
\centering

\begin{minipage}{\textwidth}
\begin{tabular}{l|rrrrrrrrrr||r}

 & \rotatebox{-45}{score} &  \rotatebox{-45}{\%gram.} &  \rotatebox{-45}{\#dif.wds.} &  \rotatebox{-45}{\#unq.wds.} &  \rotatebox{-45}{\%l.d.t.} &  \rotatebox{-45}{\#unq.n.cks.} &  \rotatebox{-45}{\#unq.q.m.a.} &  \rotatebox{-45}{fl.-kinc.} &  \rotatebox{-45}{w2v} & \rotatebox{-45}{av.s.ln.} &  \rotatebox{-45}{holistic} \\
\midrule
\multirow{5}{*}{\textbf{Lng.}} & gen. lin. & 0.695                   & 0.584                       & 0.514                   & 0.400                 &  0.493                       &  0.527                                            &    0.259                                        &    0.258            &     0.143           &    0.765    \\
\cline{2-12}
& gramm.    &   \textbf{0.698} &                      0.505 &                   0.469 &                  0.370 &                         0.423 &                                              0.468 &                             0.189   &                0.265 &                 0.134 &           0.737 \\
& orth.     &      \textbf{0.718} &                      0.395 &                   0.317 &                  0.244 &                         0.291 &                                              0.350 &                             0.155   &                0.206 &                 0.073 &           0.652 \\
& voc. ctrl.       &   0.652 &                      \textbf{0.638} &                   \textbf{0.580} &                  \textbf{0.445} &                         0.537 &                                              \textbf{0.600} &                             \textbf{0.263}  &                0.291 &                 \textbf{0.189} &           0.779 \\
& voc. rg.         &   0.651 &                      \textbf{0.621} &                   0.568 &                  0.424 &                         \textbf{0.548} &                                              0.576 &                             0.254  &                \textbf{0.339} &                 0.177 &           0.749 \\
\hline
\multirow{4}{*}{\textbf{Prg.}} & propos.  &     0.601 &                      0.607 &                   0.545 &                  0.389 &                         0.528 &                                              0.568 &                             \textbf{0.294}  &                \textbf{0.351} &                 \textbf{0.202} &           0.702 \\
& coh.   &       0.662 &                      \textbf{0.621} &                   \textbf{0.574} &                  0.410 &                         \textbf{0.551} &                                              \textbf{0.588} &                             0.248  &                0.336 &                 0.180 &           0.774 \\
& flexib.             &  0.424 &                      0.414 &                   0.390 &                  0.291 &                         0.367 &                                              0.412 &                             0.178  &                0.195 &                 0.125 &           0.443 \\
& themat.     &     0.584 &                      0.544 &                   0.527 &                  \textbf{0.428} &                         0.516 &                                              0.534 &                             0.203  &                0.287 &                 0.145 &           0.650 \\
\hline
\hline
& holistic          &     0.732 &                      0.640 &                   0.665 &                  0.451 &                         0.623 &                                              0.637 &                             0.178  &                0.364 &                 0.141 &           1.000 \\

\bottomrule
\end{tabular}
\caption{SRC correlation of the GPT-4 predicted scores and relevant linguistic features (\textbf{using holistic scores predicted by the Longformer-based grader}). The \emph{holistic} entry refers to the ground-truth holistic scores. In bold the two highest correlations columnwise.}
\label{gpt4results_predicted}
\end{minipage}

\vspace{\floatsep} 

\begin{minipage}{\textwidth}
\begin{tabular}{l|rrrrrrrrrr||r}
 & \rotatebox{-45}{score} &  \rotatebox{-45}{\%gram.} &  \rotatebox{-45}{\#dif.wds.} &  \rotatebox{-45}{\#unq.wds.} &  \rotatebox{-45}{\%l.d.t.} &  \rotatebox{-45}{\#unq.n.cks.} &  \rotatebox{-45}{\#unq.q.m.a.} &  \rotatebox{-45}{fl.-kinc.} &  \rotatebox{-45}{w2v} & \rotatebox{-45}{av.s.ln.} &  \rotatebox{-45}{holistic} \\
\midrule
\multirow{5}{*}{\textbf{Lng.}} & gen. lin. &                   0.726  &                      0.574 &                   0.541  &                  0.414 &                         0.522 &                                              0.519 &                             0.197 &                0.267 & 0.129 &           0.814 \\
\cline{2-12}
& gramm.     &                   \textbf{0.731}  &                      0.472 &                   0.464  &                  0.363 &                         0.433 &                                              0.450 &                             0.100 &                0.286 & 0.030 &          0.791 \\
& orth.     &                   \textbf{0.726}  &                      0.436 &                   0.398  &                  0.310 &                         0.354 &                                              0.427 &                             0.146 &                0.203 & 0.060 &           0.729 \\
& voc. ctrl.       &                   0.674 &                      \textbf{0.640} &                   \textbf{0.621}  &                  \textbf{0.453} &                         \textbf{0.591} &                                              \textbf{0.624} &                             \textbf{0.243} &                0.319 & 0.179 &          0.854 \\
& voc. rg.         &                   0.672  &                      \textbf{0.624} &                   \textbf{0.582} &                  \textbf{0.452} &                         \textbf{0.563} &                                              0.573 &                             0.218 &                0.280 & 0.134 &           0.816 \\
\hline
\multirow{4}{*}{\textbf{Prg.}} & propos.  &                   0.600  &                      \textbf{0.624} &                   0.581  &                  0.417 &                         0.560 &                                              \textbf{0.593} &                             \textbf{0.261} &                \textbf{0.353} & \textbf{0.190} &           0.771 \\
& coh.   &                   0.702  &                      0.555 &                   0.534  &                  0.372 &                         0.511 &                                              0.535 &                             0.238 &                \textbf{0.339} & \textbf{0.201} &           0.827 \\
& flexib.             &                   0.425  &                      0.370 &                   0.368  &                  0.249 &                         0.357 &                                              0.368 &                             0.140 &                0.163 & 0.104 &           0.488 \\
& themat.     &                   0.639  &                      0.514 &                   0.504  &                  0.413 &                         0.492 &                                              0.483 &                             0.224 &                0.264 & 0.179 &          0.745 \\
\hline
\hline
& holistic           &                   0.732  &                      0.640 &                   0.665  &                  0.451 &                         0.623 &                                              0.637 &                             0.178 &                0.364 & 0.141 &          1.000 \\

\bottomrule


\end{tabular}
\caption{SRC correlation of the GPT-4 predicted scores and relevant linguistic features (\textbf{using ground truth holistic scores}). The \emph{holistic} entry refers to the ground-truth holistic scores. In bold the two highest correlations columnwise.}
\label{gpt4results_ground}
\end{minipage}

\end{table*}

We can now move on to discussing the results of analytic scoring. Table \ref{gpt4results_predicted} shows the correlation results in terms of SRC between the predicted analytic scores and several relevant features for each proficiency aspect. Table \ref{gpt4results_ground} does the same but giving GPT-4 the ground truth holistic scores instead of the scores predicted by the holistic grader. Particularly in the latter, when focusing on the results highlighted in bold, we can observe a broad trend towards an approximate diagonal which passes through most of the proficiency aspects of the linguistic (Lng.) and pragmatic (Prg.) competences on the y-axis and the relevant features on the x-axis. For completeness, in Table \ref{gpt4results_no_score} (see Appendix \ref{sec:appendix_d}), we also report the results obtained without giving GPT-4 the holistic score, but the correlations are not as significant as the ones shown in Tables \ref{gpt4results_predicted} and \ref{gpt4results_ground} as the holistic score seems to work as a guide for analytic scoring. Furthermore, as expected, the correlations between each individual predicted analytic score and the holistic scores are significantly lower than the ones reported in Tables \ref{gpt4results_predicted} and \ref{gpt4results_ground}. Therefore, our analysis in the following lines will not dwell on these results.

As expected, grammatical error rate (\%gram.) shows the highest correlations with the aspects of grammatical accuracy and orthographic control both on Tables \ref{gpt4results_predicted} and \ref{gpt4results_ground}.

The number of unique difficult words (\#dif.wds.) seems to be a suitable feature to measure vocabulary control, e.g., if we compare the A2 level (i.e., ``Can control a narrow repertoire dealing with concrete, everyday needs.'') and the C1 level (i.e., ``Uses less common vocabulary idiomatically and appropriately.''), as described in \citet[pp. 132-133]{cefr2020} (see Appendix \ref{sec:appendix_a}). Indeed, this feature shows the highest correlation with the score related to vocabulary control.

If we look at the results obtained giving the ground truth holistic scores to GPT-4 shown in Table \ref{gpt4results_ground}, we can see that the number of unique words (\#unq.wds.), the percentage of lexical density types (\%l.d.t.), and the number of unique noun chunks (\#unq.n.cks.), which are all related to lexicality, have their highest correlation with the two scores related to vocabulary. As expected, the same features have slightly weaker --- but still relevant --- correlations when we use the automatically predicted holistic scores, as shown in Table \ref{gpt4results_predicted}.

The number of unique qualifiers, modality markers, and ambiguity indicators (\#unq.q.m.a.) is supposed to be a measure for propositional precision since, for example, as shown in Appendix \ref{sec:appendix_a}, a C1-level learner ``[c]an qualify opinions and statements precisely in relation to degrees of, for example, certainty/uncertainty, belief/doubt, likelihood, etc'' and ``[c]an make effective use of linguistic modality to signal the strength of a claim, an argument or a position'', and a C2-level learner ``[c]an convey finer shades of meaning precisely by using, with reasonable accuracy, a wide range of
qualifying devices [...]'' and ``[c]an give emphasis, differentiate and eliminate ambiguity''~\cite[p. 141]{cefr2020}. As can be observed in Table \ref{gpt4results_ground}, this feature has the second-highest correlation with the propositional precision score and the highest correlation with the score related to vocabulary control, with which it is in fact connected. Similarly to what we observed about the lexical features, the results of the fully-automated pipeline for this feature are less evident, but we can still see a rather high correlation with propositional precision.

Given its emphasis on precision and clarity, we thought that also the Flesch-Kincaid readability score (fl.kinc.) would be a suitable feature to measure these. We found that the highest correlation was exactly with propositional precision followed by vocabulary control on both Tables \ref{gpt4results_predicted} and \ref{gpt4results_ground}.

Furthermore, we considered two features for the pragmatic competence, especially in relation to cohesion and coherence. The first one is the average word2vec similarity score between all adjacent paragraphs (w2v), which shows the highest correlations on propositional precision and cohesion and coherence in Table \ref{gpt4results_ground}. The second is average sentence length (av.s.ln.), which should be an indicator of higher use of subordination and cohesive devices (i.e., longer sentences should generally be more complex). This feature shows similar results, as shown in Table \ref{gpt4results_ground}. When using the scores provided by the automatic holistic grader, the results on both features are also slightly weaker (see Table \ref{gpt4results_predicted}), as observed already for other features above.

It is rather difficult to provide a precise and exhaustive explanation of the results for the general linguistic range score, which is a broad indicator by definition since it includes elements of grammatical accuracy, syntactic complexity, and vocabulary, and, as a result, shows strong correlations with multiple features. On the other hand, the aspect of flexibility seems to be a little problematic with respect to both the features and the holistic score, probably also due to its ``longitudinality'', since it seems to be evaluated in relation to previous performances, according to its descriptors (see Appendix \ref{sec:appendix_a}).

Finally, we selected some essays in which there was a large discrepancy between two or more analytic scores, and we evaluated them impressionistically. One example can be found in Appendix \ref{sec:appendix_b}. If we focus on the highest and lowest scores, we notice vocabulary range and orthographic control on one hand, and coherence and cohesion on the other hand. Although quite extreme, this discrepancy makes sense, considering that the learner uses almost no connectors at all and mostly uses coordinating clauses (or even parataxis), but has quite a rich vocabulary and makes no orthographic errors (except for punctuation).

\subsection{Statistical tests}

\begin{figure}[ht]
    \centering
    \begin{subfigure}[b]{0.482\textwidth}
        \centering
        \includegraphics[width=\textwidth]{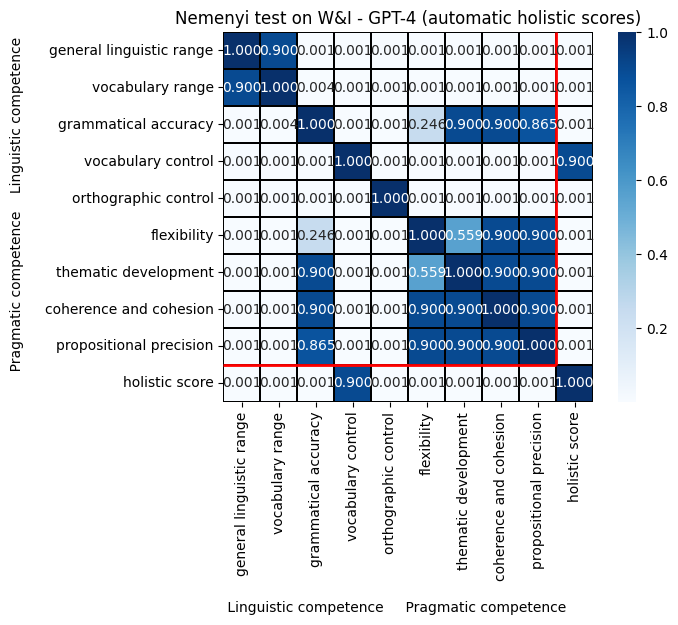}
        \label{fig:sub1}
    \end{subfigure}
    \hfill
    \begin{subfigure}[b]{0.482\textwidth}
        \centering
        \includegraphics[width=\textwidth]{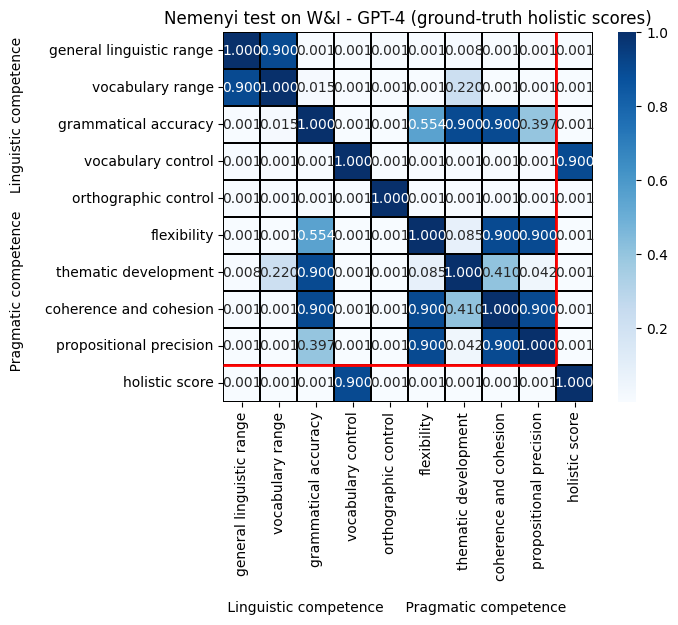}
        \label{fig:sub2}
    \end{subfigure}
    \caption{Results of the post-hoc Nemenyi test.}
    \label{nemenyi_test}
\end{figure}

Additionally, we explore the relationships among analytic scores using a repeated measures design in order to assess whether there are significant differences among them. While the repeated measures analysis of variance (rANOVA) is a widely known approach for such designs, our data fail to meet the assumptions of sphericity and normality required for its application. Hence, we employ the Friedman test~\cite{friedman}, known as the non-parametric equivalent of rANOVA. This test assesses whether there are significant differences in ranks among multiple paired groups. With a significant \emph{p}-value obtained, we confirm significant differences among the analytic scores. To determine which scores show significant differences, we conduct post-hoc multiple comparisons using the Nemenyi test~\cite{nemenyi}, whose results are reported in Figure \ref{nemenyi_test}. 
The majority of the paired comparisons, even those with the holistic
score (except when paired with vocabulary control), show significant differences (i.e., \emph{p}-value<0.05) both when we provide the ground truth and the automatic holistic scores to GPT-4. In addition to the pairs ``general linguistic range - vocabulary range'' and ``thematic development - vocabulary range'', which have some clear overlaps in their descriptors, there seem be non-significant differences over the group of aspects related to the pragmatic competence (i.e., flexibility, thematic development, coherence and cohesion, and propositional precision) and the aspect of grammatical accuracy. While we could expect to see non-significant differences among the aspects related to the pragmatic competence due to their frequent overlaps, the non-significant differences of these with grammatical accuracy might be explained with the fact that not only do its descriptors stress the importance of correctness but, as shown in Appendix \ref{sec:appendix_a}, they also emphasise complexity (e.g., for A1: ``Shows only limited control of a few simple grammatical structures [...]''; for B2: ``Has a good command of simple language structures and some complex grammatical forms [...]''), which is inherently connected to aspects such as thematic development and coherence and cohesion~\cite{purpura2004assessing}. In this regard, it is also worth noting that the coherence and cohesion score is the third most correlated with grammatical error rate.

To sum up, under ideal conditions, GPT-4 appears to produce analytic scores that are very reasonably related to the proficiency aspects they are expected to evaluate. The fully-automated pipeline is not always consistent with the ideal system but generates results that are mostly in line with it. This is especially evident for the scores pertaining to grammar and vocabulary.

\section{Conclusions}

In this paper, we have conducted an initial study on the use of GPT-4 for assessing 9 individual aspects of L2 writing underlying the CEFR proficiency levels in a zero-shot fashion. To do this, we used a holistic grading system on the essays of the W\&I validation set and, subsequently, fed them with their respective holistic scores into GPT-4, asking to assess one individual aspect at a time. Although the ground truth analytic scores are not available, we have obtained significant correlations between the predicted analytic scores and various features linked to the componential aspects of the CEFR levels. Beyond its immediate implications for computer-assisted language learning applications, we believe that our exploratory experiments may hold promise as valuable contributions to theoretical studies on construct validity in the broader field of language testing and assessment, given the inclusion of CEFR descriptors in our study.

In order to collect further evidence to support our findings, we plan to deploy this system, use it in educational settings, and evaluate its effectiveness by monitoring learners' progress in relation to each specific aspect of proficiency. Future work will also explore the use of multi-modal systems, such as the one presented in \citet{tang2023salmonn}, for assessing L2 speech in a similar fashion.

\clearpage
\section*{Limitations}

The main limitation of this study is clearly the lack of ground truth analytic scores. The reader should keep in mind, however, that, as mentioned in Section \ref{section2}, human analytic scoring is often an extremely difficult process, which might not produce completely reliable information. As evidence of this, the absence of publicly available L2 English learner datasets annotated with analytic scores speaks loud and clear and is not only an issue for the objectives of this paper, but for the whole scientific community.

\section*{Acknowledgements}

This paper reports on research supported by Cambridge University Press \& Assessment, a department of The Chancellor, Masters, and Scholars of the University of Cambridge. The authors would like to thank the ALTA Spoken Language Processing Technology Project Team for general discussions and contributions to the evaluation infrastructure.

\bibliography{custom}

\appendix

\section{Appendix A}
\label{sec:appendix_a}
\subsection*{LINGUISTIC COMPETENCE}

\subsubsection*{General linguistic range}

\noindent \textbf{A1:} Has a very basic range of simple expressions about personal details and needs of a concrete type. Can use some basic structures in one-clause sentences with some omission or reduction of elements.

\noindent \textbf{A2:} Has a repertoire of basic language which enables them to deal with everyday situations with predictable content, though they will generally have to compromise the message and search for words/signs. Can produce brief, everyday expressions in order to satisfy simple needs of a concrete type (e.g. personal details, daily routines, wants and needs, requests for information). Can use basic sentence patterns and communicate with memorised phrases, groups of a few words/signs and formulae about themselves and other people, what they do, places, possessions, etc. Has a limited repertoire of short, memorised phrases covering predictable survival situations; frequent breakdowns and misunderstandings occur in non-routine situations.

\noindent \textbf{B1:} Has a sufficient range of language to describe unpredictable situations, explain the main points in an idea or problem with reasonable precision and express thoughts on abstract or cultural topics such as music and film. Has enough language to get by, with sufficient vocabulary to express themselves with some hesitation and circumlocutions on topics such as family, hobbies and interests, work, travel and current events, but lexical limitations cause repetition and even difficulty with formulation at times.

\noindent \textbf{B2:} Can express themselves clearly without much sign of having to restrict what they want to say. Has a sufficient range of language to be able to give clear descriptions, express viewpoints and develop arguments without much conspicuous searching for words/signs, using some complex sentence forms to do so.

\noindent \textbf{C1:} Can use a broad range of complex grammatical structures appropriately and with considerable flexibility. Can select an appropriate formulation from a broad range of language to express themselves clearly, without having to restrict what they want to say.

\noindent \textbf{C2:} Can exploit a comprehensive and reliable mastery of a very wide range of language to formulate thoughts precisely, give emphasis, differentiate and eliminate ambiguity. No signs of having to restrict what they want to say.

\subsubsection*{Vocabulary range}

\noindent \textbf{A1:} Has a basic vocabulary repertoire of words/signs and phrases related to particular concrete situations.

\noindent \textbf{A2:} Has sufficient vocabulary to conduct routine everyday transactions involving familiar situations and topics. Has sufficient vocabulary for the expression of basic communicative needs. Has sufficient vocabulary for coping with simple survival needs.

\noindent \textbf{B1:} Has a good range of vocabulary related to familiar topics and everyday situations. Has sufficient vocabulary to express themselves with some circumlocutions on most topics pertinent to their everyday life such as family, hobbies and interests, work, travel and current events.

\noindent \textbf{B2:} Can understand and use the main technical terminology of their field, when discussing their area of specialisation with other specialists. Has a good range of vocabulary for matters connected to their field and most general topics. Can vary formulation to avoid frequent repetition, but lexical gaps can still cause hesitation and circumlocution. Can produce appropriate collocations of many words/signs in most contexts fairly systematically. Can understand and use much of the specialist vocabulary of their field but has problems with specialist terminology outside it.

\noindent \textbf{C1:} Has a good command of a broad lexical repertoire allowing gaps to be readily overcome with circumlocutions; little obvious searching for expressions or avoidance strategies. Can select from several vocabulary options in almost all situations by exploiting synonyms of even words/ signs less commonly encountered. Has a good command of common idiomatic expressions and colloquialisms; can play with words/signs fairly well. Can understand and use appropriately the range of technical vocabulary and idiomatic expressions common to their area of specialisation.

\noindent \textbf{C2:} Has a good command of a very broad lexical repertoire including idiomatic expressions and colloquialisms; shows awareness of connotative levels of meaning.

\subsubsection*{Grammatical accuracy}

\noindent \textbf{A1:} Shows only limited control of a few simple grammatical structures and sentence patterns in a learnt repertoire.

\noindent \textbf{A2:} Uses some simple structures correctly, but still systematically makes basic mistakes; nevertheless, it is usually clear what they are trying to say.

\noindent \textbf{B1:} Communicates with reasonable accuracy in familiar contexts; generally good control, though with noticeable mother-tongue influence. Errors occur, but it is clear what they are trying to express. Uses reasonably accurately a repertoire of frequently used “routines” and patterns associated with more predictable situations.

\noindent \textbf{B2:} Good grammatical control; occasional “slips” or non-systematic errors and minor flaws in sentence structure may still occur, but they are rare and can often be corrected in retrospect. Shows a relatively high degree of grammatical control. Does not make mistakes which lead to misunderstanding. Has a good command of simple language structures and some complex grammatical forms, although they tend to use complex structures rigidly with some inaccuracy.

\noindent \textbf{C1:} Consistently maintains a high degree of grammatical accuracy; errors are rare and difficult to spot.

\noindent \textbf{C2:} Maintains consistent grammatical control of complex language, even while attention is otherwise engaged (e.g. in forward planning, in monitoring others’ reactions).

\subsubsection*{Vocabulary control}

\noindent \textbf{A1:} No descriptors available.

\noindent \textbf{A2:} Can control a narrow repertoire dealing with concrete, everyday needs.

\noindent \textbf{B1:} Shows good control of elementary vocabulary but major errors still occur when expressing more complex thoughts or handling unfamiliar topics and situations. Uses a wide range of simple vocabulary appropriately when discussing familiar topics.

\noindent \textbf{B2:} Lexical accuracy is generally high, though some confusion and incorrect word/sign choice does occur without hindering communication.

\noindent \textbf{C1:} Uses less common vocabulary idiomatically and appropriately. Occasional minor slips, but no significant vocabulary errors.

\noindent \textbf{C2:} Consistently correct and appropriate use of vocabulary.

\subsubsection*{Orthographic control}

\noindent \textbf{A1:} Can copy familiar words and short phrases, e.g. simple signs or instructions, names of everyday objects, names of shops, and set phrases used regularly. Can spell their address, nationality and other personal details. Can use basic punctuation (e.g. full stops, question marks).

\noindent \textbf{A2:} Can copy short sentences on everyday subjects, e.g. directions on how to get somewhere. Can write with reasonable phonetic accuracy (but not necessarily fully standard spelling) short words that are in their oral vocabulary.

\noindent \textbf{B1:} Can produce continuous writing which is generally intelligible throughout. Spelling, punctuation and layout are accurate enough to be followed most of the time.

\noindent \textbf{B2:} Can produce clearly intelligible, continuous writing which follows standard layout and paragraphing conventions. Spelling and punctuation are reasonably accurate but may show signs of mother-tongue influence.

\noindent \textbf{C1:} Layout, paragraphing and punctuation are consistent and helpful. Spelling is accurate, apart from occasional slips of the pen.

\noindent \textbf{C2:} Writing is orthographically free of error.

\subsection*{PRAGMATIC COMPETENCE}

\subsubsection*{Flexibility}

\noindent \textbf{A1:} No descriptors available.

\noindent \textbf{A2:} Can adapt well-rehearsed, memorised, simple phrases to particular circumstances through limited lexical substitution. Can expand learnt phrases through simple recombinations of their elements.

\noindent \textbf{B1:} Can adapt their expression to deal with less routine, even difficult, situations. Can exploit a wide range of simple language flexibly to express much of what they want.

\noindent \textbf{B2:} Can adjust what they say and the means of expressing it to the situation and the recipient and adopt a level of formality appropriate to the circumstances. Can adjust to the changes of direction, style and emphasis normally found in conversation. Can vary formulation of what they want to say. Can reformulate an idea to emphasise or explain a point.

\noindent \textbf{C1:} Can make a positive impact on an intended audience by effectively varying style of expression and sentence length, use of advanced vocabulary and word order. Can modify their expression to express degrees of commitment or hesitation, confidence or uncertainty.

\noindent \textbf{C2:} Shows great flexibility in reformulating ideas in differing linguistic forms to give emphasis, differentiate according to the situation, interlocutor, etc. and to eliminate ambiguity.

\subsubsection*{Thematic development}

\noindent \textbf{A1:} No descriptors available.

\noindent \textbf{A2:} Can tell a story or describe something in a simple list of points. Can give an example of something in a very simple text using “like” or “for example”.

\noindent \textbf{B1:} Can clearly signal chronological sequence in narrative text. Can develop an argument well enough to be followed without difficulty most of the time. Shows awareness of the conventional structure of the text type concerned when communicating their ideas. Can reasonably fluently relate a straightforward narrative or description as a sequence of points.

\noindent \textbf{B2:} Can develop an argument systematically with appropriate highlighting of significant points, and relevant supporting detail. Can present and respond to complex lines of argument convincingly. Can follow the conventional structure of the communicative task concerned when communicating their ideas. Can develop a clear description or narrative, expanding and supporting their main points with relevant supporting detail and examples. Can develop a clear argument, expanding and supporting their points of view at some length with subsidiary points and relevant examples. Can evaluate the advantages and disadvantages of various options. Can clearly signal the difference between fact and opinion.

\noindent \textbf{C1:} Can use the conventions of the type of text concerned to hold the target reader’s attention and communicate complex ideas. Can give elaborate descriptions and narratives, integrating sub-themes, developing particular points and rounding off with an appropriate conclusion. Can write a suitable introduction and conclusion to a long, complex text. Can expand and support the main points at some length with subsidiary points, reasons and relevant examples.

\noindent \textbf{C2:} Can use the conventions of the type of text concerned with sufficient flexibility to communicate complex ideas in an effective way, holding the target reader’s attention with ease and fulfilling all communicative purposes.

\subsubsection*{Propositional precision}

\noindent \textbf{A1:} Can communicate basic information about personal details and needs of a concrete type in a simple way.

\noindent \textbf{A2:} Can communicate what they want to say in a simple and direct exchange of limited information on familiar and routine matters, but in other situations they generally have to compromise the message.

\noindent \textbf{B1:} Can explain the main points in an idea or problem with reasonable precision. Can convey simple, straightforward information of immediate relevance, getting across the point they feel is most important. Can express the main point they want to make comprehensibly.

\noindent \textbf{B2:} Can pass on detailed information reliably. Can communicate the essential points even in more demanding situations, though their language lacks expressive power and idiomaticity.

\noindent \textbf{C1:} Can qualify opinions and statements precisely in relation to degrees of, for example, certainty/uncertainty, belief/doubt, likelihood, etc. Can make effective use of linguistic modality to signal the strength of a claim, an argument or a position.

\noindent \textbf{C2:} Can convey finer shades of meaning precisely by using, with reasonable accuracy, a wide range of qualifying devices (e.g. adverbs expressing degree, clauses expressing limitations). Can give emphasis, differentiate and eliminate ambiguity.

\subsubsection*{Coherence and cohesion}

\noindent \textbf{A1: } Can link words/signs or groups of words/signs with very basic linear connectors (e.g. “and” or “then”).

\noindent \textbf{A2: } Can use the most frequently occurring connectors to link simple sentences in order to tell a story or describe something as a simple list of points. Can link groups of words/signs with simple connectors (e.g. “and”, “but” and “because”).

\noindent \textbf{B1: } Can introduce a counter-argument in a simple discursive text (e.g. with “however”). Can link a series of shorter, discrete simple elements into a connected, linear sequence of points. Can form longer sentences and link them together using a limited number of cohesive devices, e.g. in a story. Can make simple, logical paragraph breaks in a longer text.

\noindent \textbf{B2: } Can use a variety of linking expressions efficiently to mark clearly the relationships between ideas. Can use a limited number of cohesive devices to link their utterances into clear, coherent discourse, though there may be some “jumpiness” in a long contribution. Can produce text that is generally well-organised and coherent, using a range of linking expressions and cohesive devices. Can structure longer texts in clear, logical paragraphs.

\noindent \textbf{C1: } Can produce clear, smoothly flowing, well-structured language, showing controlled use of organisational patterns, connectors and cohesive devices. Can produce well-organised, coherent text, using a variety of cohesive devices and organisational patterns.

\noindent \textbf{C2: } Can create coherent and cohesive text making full and appropriate use of a variety of organisational patterns and a wide range of cohesive devices.

\section{Appendix B}
\label{sec:appendix_b}

\emph{I deal with consulting and sales of financial products and services to an international bank, in the mass-market and small-business. I follow the relationship with customers from acquisition to the advise until the realization of contracts, building and maintaining relationships after-sales in the aim of customer satisfaction}

\emph{I also worked with large and small teams in back-offices, managed many administrative activities related to mortages, personal loans, contability and investments too.}

\emph{I worked for several years to the acquisition of new customers, to provide them with a complete service, from the account to insurance products, investment products, personal loans, revolving credit, and cross-selling products. In many years of work I have honed my skills in managing  non-standard situations, analyzing the problem, finding and implementing practical and easy solutions. non-standard situations, analyzing the problem, finding and implementing practical and easy solutions.}

\emph{I have faced several situations always work with serenity and enthusiasm, I like to work in a multicultural and dynamic.}

\emph{I'm careful to meet the goals of the team in which I work, cooperating with colleagues to achieve the goals by providing my skills, always willing to learn, respecting other points of view together finding ways to deal. I work for the same large company for 25 years, now is the time to change and find new job opportunities. Needs to work my husband has been living in Zaandam, I want to find a new job in Holland to rejoin our family.}

\emph{I like sports such as skiing, riding and swimming. I've also got the rescue licence, I worked as a lifeguard in the summer studying for the patent padi dive master}

The holistic score is 3.5 (B1+), and GPT-4 provided these analytic scores:
\begin{itemize}
    \item general linguistic range: 3
    \item vocabulary range: 4
    \item grammatical accuracy: 2
    \item vocabulary control: 3
    \item orthographic control: 4
    \item flexibility: 2
    \item thematic development: 2
    \item coherence and cohesion: 1
    \item propositional precision: 3
\end{itemize}

\section{Appendix C}
\label{sec:appendix_c}

When we include the holistic score, the prompt given to GPT-4 is the following:

\begin{quote}

 \texttt{Consider the following essay: [\emph{ESSAY}]}

 \texttt{It has been given this score on a scale from 1 to 6.5: [\emph{HOLISTIC SCORE}].}

 \texttt{I want you to assess it only considering the aspect of [\emph{ASPECT}], for which you have 6 different feedback options, that you will have to accept or reject: [\emph{ANALYTIC CEFR DESCRIPTORS}]}
 
 \texttt{ONLY ONE option can be accepted and is the option you will have to output by only selecting the option letter in the following format: 'option A/B/C/D/E/F'\footnote{\label{footnote_options}The aspects of vocabulary control, flexibility, and thematic development only have options A-E since no descriptors are available for the A1 level.} WITHOUT ANY ADDITIONAL OBSERVATION, COMMENT, NOTE, EXPLANATION, CLARIFICATION, OR JUSTIFICATION OF ANY SORT.}

\texttt{Your answer:}
\end{quote}

When we do not provide GPT-4 with the holistic score, the prompt is the following:

\begin{quote}

 \texttt{Consider the following essay: [\emph{ESSAY}]}

 \texttt{I want you to assess it only considering the aspect of [\emph{ASPECT}], for which you have 6 different feedback options, that you will have to accept or reject: [\emph{ANALYTIC CEFR DESCRIPTORS}]}
 
 \texttt{ONLY ONE option can be accepted and is the option you will have to output by only selecting the option letter in the following format: 'option A/B/C/D/E/F'\footnote{See note \ref{footnote_options}.} WITHOUT ANY ADDITIONAL OBSERVATION, COMMENT, NOTE, EXPLANATION, CLARIFICATION, OR JUSTIFICATION OF ANY SORT.}

\texttt{Your answer:}
\end{quote}

\section{Appendix D}
\label{sec:appendix_d}

\subsubsection*{Score alignment}

Table \ref{scores_alignment} shows the holistic score normalisation process for EFCAMDAT.

\begin{table}[ht]

\centering
\begin{tabular}{c|c|c}
\hline
\textbf{CEFR} & \textbf{W\&I} & \textbf{EFCAMDAT}  \\
\hline
\multirow{2}{*}{A1}                       & A1 (1)        & 1,2     \\
                               &    A1+ (1.5)     & 3     \\
                               \hline
\multirow{2}{*}{A2}                       & A2 (2)        & 4,5     \\
                               &    A2+   (2.5)   & 6     \\
                               \hline
\multirow{2}{*}{B1}                       & B1 (3)        & 7,8     \\
                               &    B1+  (3.5)    & 9     \\
                               \hline
\multirow{2}{*}{B2}                       & B2   (4)      & 10,11     \\
                               &    B2+  (4.5)    & 12     \\
                               \hline
\multirow{2}{*}{C1}                       & C1     (5)    & 13,14     \\
                               &    C1+  (5.5)    & 15     \\
                               \hline
\multirow{2}{*}{C2}                       & C2   (6)      & 16 (score$<$85)     \\
                               &    C2+  (6.5)    & 16 (score$\geq$85)     \\
                               \hline

\hline
\end{tabular}
\caption{Score alignment.}
\label{scores_alignment}
\end{table}

\subsubsection*{Additional experimental results}

Table \ref{gpt4results_no_score} reports the results of the experiment conducted when no holistic scores are given to GPT-4.

\begin{table*}[ht!]
\footnotesize
\centering

\begin{minipage}{\textwidth}
\begin{tabular}{l|rrrrrrrrrr||r}

 & \rotatebox{-45}{score} &  \rotatebox{-45}{\%gram.} &  \rotatebox{-45}{\#dif.wds.} &  \rotatebox{-45}{\#unq.wds.} &  \rotatebox{-45}{\%l.d.t.} &  \rotatebox{-45}{\#unq.n.cks.} &  \rotatebox{-45}{\#unq.q.m.a.} &  \rotatebox{-45}{fl.-kinc.} &  \rotatebox{-45}{w2v} & \rotatebox{-45}{av.s.ln.} &  \rotatebox{-45}{holistic} \\
\midrule
\multirow{5}{*}{\textbf{Lng.}} & gen. lin. &                   0.643 &                      0.622 &                   0.547 &                  0.471 &                         0.526 &                                              0.565 &                             0.268 &                0.275 &                 0.148 &           0.739 \\
\cline{2-12}
& gramm.     &                   \textbf{0.707} &                      0.408 &                   0.365 &                  0.284 &                         0.324 &                                              0.364 &                             0.151 &                0.170 &                 0.099 &           0.692 \\
& orth.     &                   \textbf{0.730} &                      0.362 &                   0.290 &                  0.234 &                         0.259 &                                              0.309 &                             0.133 &                0.166 &                 0.068 &           0.653 \\
& voc. ctrl.       &                   0.697 &                      0.391 &                   0.363 &                  0.305 &                         0.331 &                                              0.369 &                             0.153 &                0.102 &                 0.107 &           0.654 \\
& voc. rg.         &                   0.529 &                      0.539 &                   0.456 &                  \textbf{0.410} &                         0.450 &                                              0.452 &                             \textbf{0.247} &                0.241 &                 0.131 &           0.616 \\
\hline
\multirow{4}{*}{\textbf{Prg.}} & propos.  &                   0.432 &                      0.510 &                   0.442 &                  0.341 &                         0.430 &                                              0.492 &                             \textbf{0.246} &                \textbf{0.304} &                 0.145 &           0.539 \\
& coh.   &                   0.602 &                      \textbf{0.601} &                   \textbf{0.542} &                  0.379 &                         \textbf{0.533} &                                              \textbf{0.571} &                             0.244 &                0.299 &                 \textbf{0.162} &           0.729 \\
& flexib.              &                   0.307 &                      0.361 &                   0.363 &                  0.282 &                         0.348 &                                              0.346 &                             0.202 &                0.149 &                 \textbf{0.160} &           0.330 \\
& themat.     &                   0.425 &                      \textbf{0.612} &                   \textbf{0.587} &                  \textbf{0.496} &                         \textbf{0.576} &                                              \textbf{0.583} &                             0.242 &                \textbf{0.333} &                 0.150 &           0.543 \\
\hline
\hline
& holistic           &                   0.732 &                      0.640 &                   0.665 &                  0.451 &                         0.623 &                                              0.637 &                             0.178 &                0.364 &                 0.141 &           1.000 \\
\bottomrule
\end{tabular}
\caption{SRC correlation of the GPT-4 predicted scores and relevant linguistic features (\textbf{without giving GPT-4 the holistic score}). The \emph{holistic} entry refers to the ground-truth holistic scores. In bold the two highest correlations columnwise.}
\label{gpt4results_no_score}
\end{minipage}

\vspace{\floatsep} 

\end{table*}

\end{document}